\ifx\pfmtname\undefined
\documentclass[a4paper]{article}
\else
\documentclass{article}
\fi
\usepackage{robomech_en}
\usepackage{graphicx}
\usepackage{amsfonts}

\begin{document}
\makeatletter
\title{Learning Bipedal Walking for Humanoid Robots in Challenging Environments with Obstacle Avoidance}
{}
{}
{}

\author{ \small
\begin{tabular}{ll}
 $\bigcirc$ & Marwan HAMZE, Tokyo University of Science, marwan.hamze@rs.tus.ac.jp\\
  & Mitsuharu MORISAWA, National Institute
of Advanced Industrial Science and Technology (AIST)\\
  & Eiichi YOSHIDA, Member, Tokyo University of Science\\
\end{tabular}
}
\makeatother

\abstract{ \small
Deep reinforcement learning has seen successful implementations on humanoid robots to achieve dynamic walking. However, these implementations have been so far successful in simple environments void of obstacles. In this paper, we aim to achieve bipedal locomotion in an environment where obstacles are present using a policy-based reinforcement learning. By adding simple distance reward terms to a state of art reward function that can achieve basic bipedal locomotion, the trained policy succeeds in navigating the robot towards the desired destination without colliding with the obstacles along the way.
}

\date{} 

\maketitle
\thispagestyle{empty}
\pagestyle{empty}

\small
\section{Introduction}

Locomotion for legged robots has been studied for decades, and continues to be a topic of research. Bipedal locomotion, especially for humanoid robots, has been a difficult topic to address, due to difficulties concerning balance, collision, and efficiency. For humanoid robots, controllers are designed in order to ensure a balanced motion for a specific robot. The most well-known controllers are basically considered model-based controllers, in the sense that the robot and the interaction with the environment are modeled according to the law of physics and mechanics. These controllers, such as the ones based on the linear inverted pendulum mode \cite{caron_stair_2019}\cite{kajita_biped_2010}, rely on the accuracy of their simplified representation of the natural interaction between the robot and the environment, which is always an imperfect representation of the complex interaction. This is why they have limited robustness in uncertain environments, even when relying on stabilizing control to counter these modeling errors. Recently however, researchers have been testing with reinforcement learning-based controllers \cite{kim_torque-based_2023} \cite{rodriguez_deepwalk_2021}. These controllers aim to train the robot in simulation to obtain an optimal policy, allowing the robot to achieve robust locomotion in uncertain environments by adapting its movement in these environments during training, without the need for precise modeling of the robot-environment interaction. However, reinforcement learning-based controllers come with their own challenges, such as their unpredictable behavior, which is driven by the trained policy. This is one reason why their application is so far limited to simple environments void of obstacles. 

In this paper, we work on executing a bipedal locomotion in a challenging environment, where obstacles exist. We implement a simple learning method for a humanoid robot in order to achieve a collision-free locomotion to arrive at a desired destination. This method is based on the actor-critic algorithm usually used in previous works for bipedal locomotion, with additional terms for the reward function to make a collision-free movement possible.

\begin{figure*}[tbh]
    \centering
    \includegraphics[height=58mm]{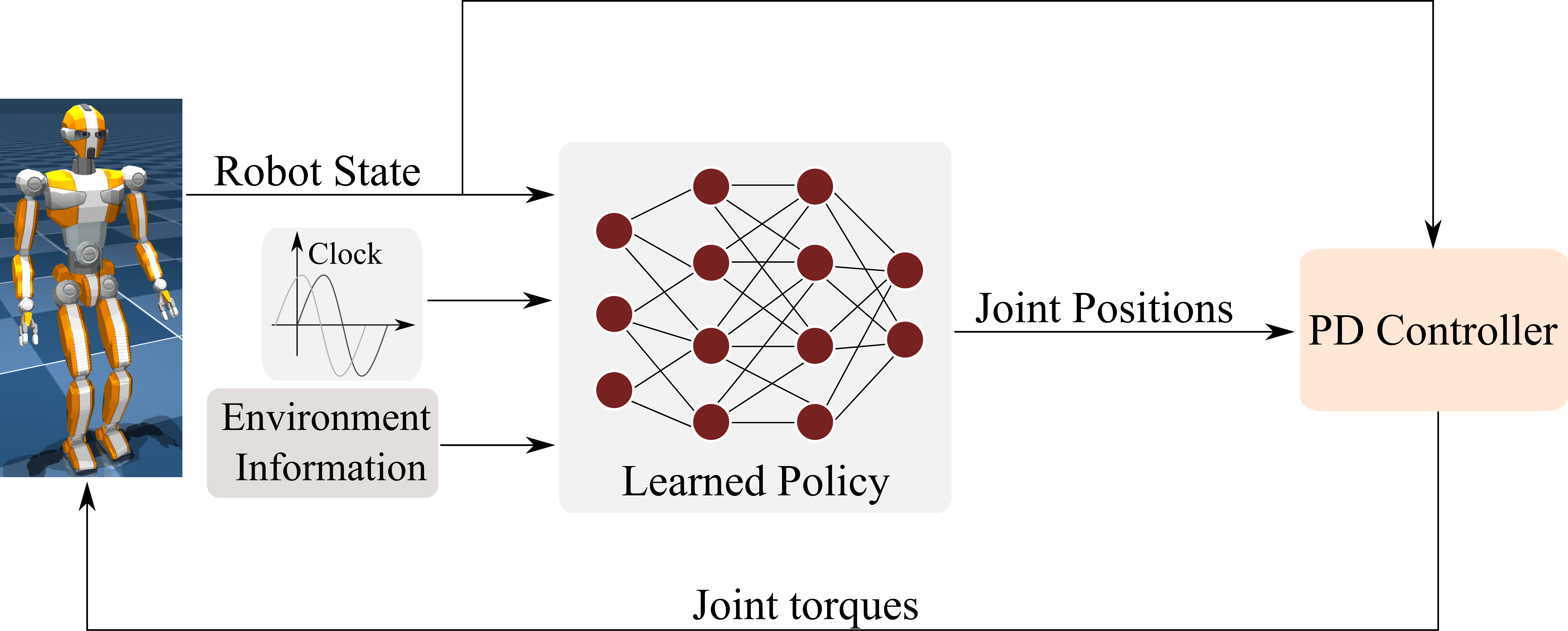}
    \vspace*{-4mm}
    \caption{Overall control structure. The environment information consists of the obstacle and destination reference positions.}
    \label{fig: diagram}
\end{figure*}

\section{Background}

Reinforcement Learning \cite{qiang_reinforcement_2011} aims to maximize a reward $r$ by prescribing an action $a$ given an input $s$. The world is represented as a discrete-time Markov Decision Process (MDP), formed by a continuous state space $S\in\mathbb{R}^{n}$ , and action space $A\in\mathbb{R}^{m}$, a state-transition function $p(s, a, s')$ and a reward function $r(s, a)$. The state-transition function, unknown $a priori$, defines the dynamics of the world and gives the probability density over the next state $s'$ when taking action $a$ in the current state $s$. The reward function provides a scalar signal at each state transition that the agent aims to maximize.
The policy $\pi\left(a\mid s\right)$ is defined as a stochastic mapping from states to action. The goal in reinforcement learning is to find a $\pi$ that maximizes the agent’s expected T-horizon discounted return given by $J\left(\pi_{\theta}\right)=\mathbb{E}\left[\sum_{t=0}^{T}\gamma^{t}r\left(s_{t},a_{t}\right)\right]$, where $\gamma\in\left(0,1\right]$ is the discount factor. In the case of large, continuous state
spaces, we use a parametric policy $\pi_{theta}$ with parameters $\theta$ , which represent the set of parameters of a multi-layered perceptron (MLP). The policy is improved iteratively, by estimating the gradient of $J\left(\pi_{\theta}\right)$ and updating the policy parameters by performing stochastic gradient ascent with a step size $\alpha: \theta_{k+1}=\theta_{k}+\alpha\nabla J\left(\pi_{\theta_{k}}\right)$. $\nabla J\left(\pi_{\theta_{k}}\right)$ can be
estimated with the experience collected from trajectories sampled by following the policy $\pi_{\theta_{k}}$.
We use the Proximal Policy Optimization (PPO) algorithm \cite{schulman_proximal_2017}, which builds upon the above-explained policy gradient method, in order to increase sample efficiency while avoiding policy collapse.

\section{Controller Design}

Consider a humanoid robot with $n$ degrees of freedom. The controller adopts a high-level policy to calculate joint positions, which are fed to low-gains PD controllers, responsible for calculating the joint torques.

\subsection{Observation and Action Spaces}%

The \textbf{Observation Space} consists of the robot's state taken from feedback measurements, reference values, and a clock signal. The robot's space includes joint positions and velocities of the robot's leg joints, in addition to the floating base's rotation and angular velocity. 

The reference values consist of the fixed obstacle positions in the environment in addition to the goal position, which is the destination where the robot is supposed to arrive. Theses references are given by their x-y coordinates, and are required for the reward terms introduced later, so that the robot is aware of the objects to avoid and of its destination.

The clock signal is necessary for generating periodic locomotion based on periodic reward terms. We basically use the same clock signal as the one introduced in \cite{singh_learning_2022}, where you can read about it in detail.

As for the \textbf{Action Space}, the policy produces the joint positions of the leg joints only, in order to facilitate the learning of bipedal locomotion. We apply stiff PD control to the rest of the joints to maintain a half-sitting posture.  

\subsection{Reward function}%

We will devise the reward function into two parts: a \textit{basic locomotion} part and \textit{distance} part, which can be written as:

\begin{equation}
r=W_{1}r_{locomotion}+W_{2}r_{distance},
\end{equation}

The basic locomotion part $r_{locomotion}$ consists of generic terms used frequently in the state of the art of bipedal locomotion in reinforcement learning. We adopt the reward terms from \cite{singh_learning_2022}, minus the reward term that is concerned with following the footsteps provided by a higher-level footstep planner, since we are not using one.

The distance part $r_{distance}$ consists of reward terms to avoid collision with the obstacles in the environment and to reach the desired destination. The reward term is based on the distance between the robot's base position and the target in question, written as:
\begin{equation}
r_{distance}=\sum_{i=1}^{m}e^{-k_{i}\left\Vert p_{base}\left(x,y\right)-p_{i,target}\left(x,y\right)\right\Vert },
\end{equation}

where $k$ is a constant that is tuned for the desired behavior, and $m$ is the number of distance rewards to be added. $p_{i,target}\left(x,y\right)$ represents the position of an obstacle, the destination, or the initial position of the floating base. We use this term as a positive reward when it is applied to the destination; this encourages the policy to minimize the distance between the robot's base and the desired destination. On the other hand, we use this term as a negative reward with each obstacle in the environment, and with the initial position of the floating base; this encourages the policy to navigate the robot away from the obstacles. We also realized that without a negative distance reward on the initial position, the robot would step in place around its initial position instead of heading towards the goal, as the policy concludes that not approaching the obstacles is more beneficial. 

$W_{1}$ is a weight matrix given to the reward terms for the basic locomotion, and $W_{2}$ is attributed to the distance-based terms. The weights for the locomotion were set as in \cite{singh_learning_2022}, while the distance-based weights require some tuning; basically we set the weight for the distance to the destination as $0.95$, while setting the one for the distance to each obstacle as $-0.2$, and as $-0.5$ for the initial position. We noticed that giving a significant weight for each obstacle hinders the walking performance, as the robot starts reducing its step size.

An overview of the control structure is given in Figure \ref{fig: diagram}. The joint positions calculated by the policy are added to the default values for a half-sitting position, before being converted to joint torques.

\section{Results}

\begin{figure}[tbh]
    \centering
    \includegraphics[height=42mm]{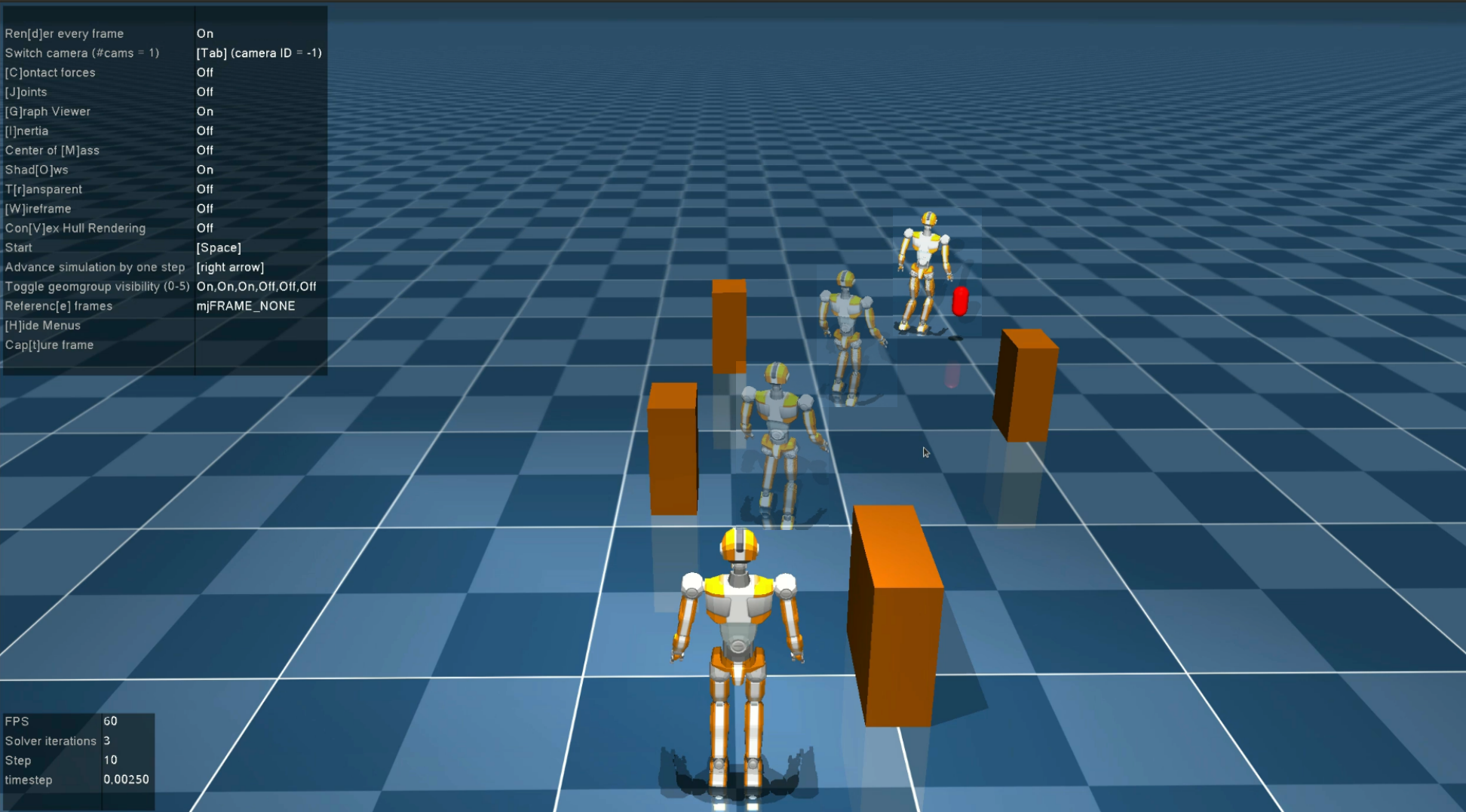}
    \vspace*{-4mm}
    \caption{The simulated environment, highlighting the trajectory taken by the robot to reach the destination in red.}
    \label{fig: mujoco}
\end{figure}

The robot used in our simulations is the JVRC-1 virtual robot model, which was developed for the Japan Virtual Robotics Challenge \cite{okugawa_proposal_2015}. This 172cm tall humanoid robot weighs 62kg and has 34 degrees of freedom (d.o.f). The robot has $6$ d.o.f in each leg, which means that the action space of the policy has a dimension of $12$.
We used the MLP architecture to represent both the actor and critic policies, which parameterize the policy and the value function in PPO \cite{schulman_proximal_2017}. Each MLP network consists of 2 hidden layers of size 256 and uses ReLU activations, before passing the output through a $tanh$ layer to limit the range of the actor’s predictions. The hyperparameters are primarily similar to \cite{singh_learning_2022}, and we set the learning episode length to $400$ iterations. The policy was trained for about 10 hours on 96 million samples, with the simulations being run entirely on an Intel Core i9-13900HXH CPU @ 2.2GHz with 32 cores. We trained the policy for different environments with different obstacles and destination, however we will talk specifically about the case represented in Figure \ref{fig: mujoco}, where there are 4 small obstacles that the robot needs to not collide with before reaching its destination marked in red.

\begin{figure}[tbh]
    \centering
    \includegraphics[height=58mm]{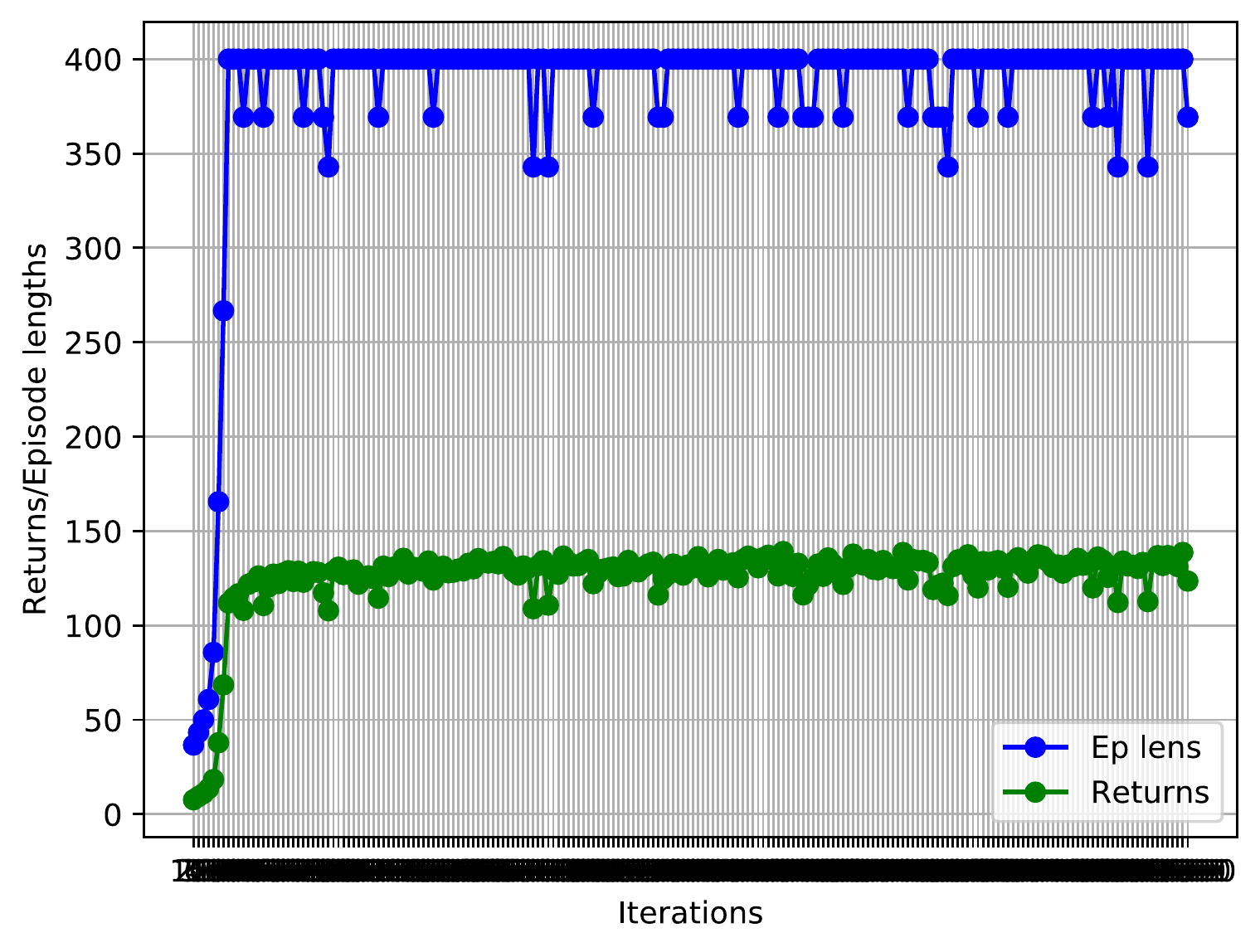}
    \vspace*{-4mm}
    \caption{Episode Lengths and Returns during the Learning Process}
    \label{fig: eval}
\end{figure}

Figure \ref{fig: eval} shows the return value and episode lengths throughout the learning process. The return value represents the total reward at each timestep during the learning process, averaging at around $138$ with little variance, meaning that the number of samples was enough to obtain a good policy. The episodes finished at their $400$ iterations most of the time, and were terminated a few times because of terminating conditions, such as self-collision or in case the robot fell during training, which are set just like in \cite{singh_learning_2022}.

\subsubsection*{}{\textbf{Walking with Obstacle Avoidance}}

Thanks to the simple distance reward terms added to the regular basic walking reward terms, the robot succeeds in its motion and reaches the destination without collisions. We noticed that the robot makes smaller steps compared to a case where the robot is trained to walk on a terrain without any obstacles. This is probably due to the increased difficulty of the executed motion, so the policy had to penalize the center of mass velocity reward. A video of the simulations is accessible in \cite{Simulations}.

\subsubsection*{}{\textbf{Robustness}}

In order to test the robustness of the learned policy, simulations were done while adding a random displacement to the position of each obstacle and to the destination. The learned policy successfully executes the desired motion when the obstacles are moved in the frontal plane, however collisions occur when they are moved along the sagittal plane for about $0.5m$. This is probably because it is more challenging for the robot to move in the sagital plane while heading for the desired destination. The robot can still arrive at the destination when it's displaced by up to $1.5m$ in the sagittal plane, and the displacement in the frontal plane causes no issues for the policy.

\subsubsection*{}{\textbf{Discussion}}

Our simple modifications to the reward function allowed the robot to safely execute a walking motion in a challenging environment without collisions. However, this approach is currently limited to fixed and known obstacle positions. While the policy demonstrated some robustness with respect to the obstacles' positions, it would not work if the obstacles were completely randomized or unknown. In such cases, it is necessary to observe the obstacle's positions using vision-based sensors (e.g., camera, LiDAR), and rely on a Visual-locomotion policy architecture such as  \cite{escontrela_zero-shot_2020}.

Another topic to mention is that the executed motion, like other related works on humanoid robots, actuates only the degrees of freedom of the robot's legs. Our research team is interested in expanding the basic locomotion into a multi-contact locomotion: the robot should exploit the presence of the objects in the environment, and use them to form additional contacts with its hands before arriving at the destination. This paper serves as a first step towards reaching this goal, by showing how the learned policy with the introduced modifications to the regular walking policy behaves in a challenging environment with objects.

\section{Conclusion}

To go a step beyond basic locomotion for humanoid robots in simple environments, our proposed adjustments to the regular reward function for bipedal walking, consisting of adding negative distance reward terms to avoid obstacles and a positive one to reach the desired destination, allow the robot to execute collision-free locomotion and arrive at the desired destination. Future work will aim to improve the reinforcement learning algorithm, to make the robot interact with the objects in the environment by executing multi-contact motion before reaching its desired destination.

\section*{Acknowledgments}
The authors would like to thank Dr. Rohan Singh for his  feedback and useful insights.

\footnotesize
\bibliographystyle{IEEEtran3etal} 
\bibliography{./Robomech}

\normalsize
\end{document}